\pdfoutput=1

\documentclass[11pt]{article}

\usepackage[preprint]{acl}

\usepackage{times}
\usepackage{latexsym}

\usepackage[T1]{fontenc}

\usepackage[utf8]{inputenc}

\usepackage{microtype}

\usepackage{inconsolata}

\usepackage{graphicx}

\usepackage{amssymb}
\usepackage{tabularx}
\usepackage{booktabs} 
\usepackage{pdfpages}
\usepackage{enumitem}
\usepackage{color, colortbl}
\usepackage{xcolor}
\usepackage{pifont}
\usepackage{multirow}
\definecolor{Red}{rgb}{1,0,0}
\definecolor{LimeGreen}{rgb}{0.2,0.8,0.2}
\usepackage{xspace}
\newcommand{\ours}{{DrugAgent}\xspace}

\usepackage{listings}
\usepackage{xcolor}

\definecolor{dkgreen}{rgb}{0,0.6,0}
\definecolor{gray}{rgb}{0.5,0.5,0.5}
\definecolor{mauve}{rgb}{0.58,0,0.82}

\usepackage{fvextra} 

\usepackage[]{mdframed}
\usepackage{tikz}

\title{\ours: Automating AI-aided Drug Discovery Programming through LLM Multi-Agent Collaboration}

\author{%
Sizhe Liu$^{1}$, \textbf{Yizhou Lu}$^{1}$, \textbf{Siyu Chen}$^{1}$, \textbf{Xiyang Hu}$^{2}$, \textbf{Jieyu Zhao}$^1$, \textbf{Yingzhou Lu}$^3$, \textbf{Yue Zhao}$^{1}$\thanks{Corresponding author}\\ 
$^1$University of Southern California \quad $^2$Carnegie Mellon University \quad $^3$Stanford University \\%
}

\begin{document}
\maketitle

\begin{abstract}
Recent progress in Large Language Models (LLMs) has drawn attention to their potential for accelerating drug discovery. 
However, a central problem remains: translating theoretical ideas into robust implementations in the highly specialized context of pharmaceutical research. 
This limitation prevents practitioners from making full use of the latest AI developments in drug discovery. 
To address this challenge, we introduce \ours, a multi-agent framework that automates machine learning (ML) programming for drug discovery tasks. 
\ours employs an \textit{LLM Planner} that formulates high-level ideas and an \textit{LLM Instructor} that identifies and integrates domain knowledge when implementing those ideas. 
We present case studies on three representative drug discovery tasks. 
Our results show that \ours consistently outperforms leading baselines, including a relative improvement of 4.92\% in ROC-AUC compared to ReAct for drug-target interaction (DTI). 
\ours is publicly available at the anonymous link \url{https://anonymous.4open.science/r/drugagent-5C42/}.
\end{abstract}

\section{Introduction and Related Work}

Artificial intelligence (AI) is changing many aspects of drug discovery \cite{huang2022artificial}. 
Since experimental measurements of drug properties are costly and time-consuming, researchers have turned to automated approaches for diverse stages of drug development~\cite{Pushpakom2019}. 
AI-ready datasets and benchmarks, such as ADMET prediction, drug-target interaction, and high-throughput screening, are now widely accessible~\cite{Huang2021TherapeuticsDC,chen2024trialbench,wang2024twin}. 
Meanwhile, deep learning has shown promise in lead optimization and drug-target interaction prediction~\cite{huang2020deeppurpose}, pointing toward possible reductions in the resources required for traditional experimentation.

Yet building machine learning (ML) pipelines for drug discovery is challenging, given that it involves biology, chemistry, pharmaceutical science, and computer science~\cite{huang2022artificial}. 
While Large Language Models (LLMs) offer automated reasoning and coding assistance, domain-specific subtleties remain difficult to handle in standard frameworks. 
General-purpose agent-based systems for ML, such as MLAgentBench~\cite{huang2024mlagentbench} and AI-Scientist~\cite{lu2024aiscientist}, have been proposed for end-to-end ML programming, but they lack expert-level knowledge of drug discovery workflows. 
Small mistakes, such as using the wrong domain-specific library or misinterpreting biological data types, can be difficult to debug in specialized projects. 
In contrast, frameworks like ChemCrow~\cite{Bran} and MultiTool-CoT  (Chain of Thought)~\cite{inaba-etal-2023-multitool} include chemical tools but offer limited support for larger-scale ML tasks. 
This highlights the need for \textit{an ML-focused system with domain awareness}, spanning data preprocessing through model evaluation.



\begin{table*}[ht]
  \centering
  \caption{\textbf{Key differences between \ours and existing agent methods.} \ours stands out by: 1) interacting with the environment, 2) specializing in ML programming, 3) incorporating domain knowledge specific to drug discovery, and 4) planning at the idea space level.}
  \vspace{-2mm}
  \label{tab:setting}
  \resizebox{\textwidth}{!}{%
  \begin{tabular}{lcccc}
    \toprule
     & Interaction with Env & ML Specialization & Domain Knowledge & Idea Space Planning \\
     \midrule
     ReAct~\cite{yao2023react}
     & \textcolor{LimeGreen}{\ding{52}} 
     & \textcolor{Red}{\ding{55}} 
     & \textcolor{Red}{\ding{55}}
     & \textcolor{Red}{\ding{55}} \\
     ResearchAgent~\cite{huang2024mlagentbench}
     & \textcolor{LimeGreen}{\ding{52}}
     & \textcolor{LimeGreen}{\ding{52}}
     & \textcolor{Red}{\ding{55}}
     & \textcolor{Red}{\ding{55}} \\
     ChemCrow~\cite{Bran}
     & \textcolor{LimeGreen}{\ding{52}}
     & \textcolor{Red}{\ding{55}}
     & \textcolor{LimeGreen}{\ding{52}}
     & \textcolor{Red}{\ding{55}} \\
     \ours (Ours)
     & \textcolor{LimeGreen}{\ding{52}}
     & \textcolor{LimeGreen}{\ding{52}}
     & \textcolor{LimeGreen}{\ding{52}}
     & \textcolor{LimeGreen}{\ding{52}} \\
    \bottomrule
  \end{tabular}
  }
  \vspace{-5pt}
\end{table*}



\begin{figure*}[t]
    \centering
    \includegraphics[width=\linewidth]{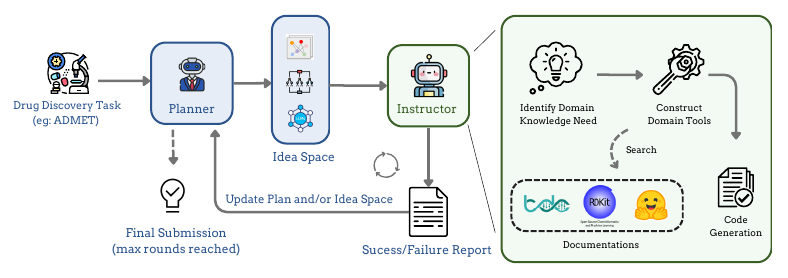}
    \vspace{-2mm}
    \caption{Overview of the \ours framework. Given a drug discovery task described in natural language (i.e., user's input, e.g., design an AI model to predict Absorption (one of the ADMET properties) using the PAMPA dataset~\cite{Siramshetty2021}), the LLM Planner collaborates with the LLM Instructor to iteratively search for actionable, high-performing solutions.}
    \label{fig:overview-agent}
\end{figure*}


\noindent\textbf{Present Work: \ours.} We introduce \ours, a multi-agent LLM framework that unifies ML programming with biomedical expertise to address the demands of modern drug discovery. 
First, \ours systematically checks where domain knowledge is required, then deploys specialized resources before proceeding with coding. 
Second, it uses a dynamic approach to manage ML ideas, creating diverse options early on and refining them based on empirical results. 
Third, \ours features a carefully curated library of domain-specific documentation covering data acquisition, data transformation, and advanced model design, supporting critical tasks in drug discovery. 
We evaluate \ours on three representative tasks and find that it exceeds the performance of general-purpose baselines and matches or surpasses expert-written methods. 
\textbf{Our key contributions include}: 
\textbf{(1)} a systematic workflow that emphasizes when and how to incorporate domain knowledge for ML-driven drug discovery, 
\textbf{(2)} an iterative planning strategy guided by experimental observations, and 
\textbf{(3)} a broad set of specialized tools and documentation for biological data processing and modeling. 
A detailed comparison with existing approaches is in Table~\ref{tab:setting} and Appendix~\ref{appendix:related_work}.


    
    

\section{Methodology}

We present \ours, a multi-agent LLM framework designed to handle the specialized challenges of AI-driven drug discovery. 
As illustrated in Figure~\ref{fig:overview-agent}, \ours integrates two primary agents: 
(1)~an LLM \textbf{Planner}, which manages the high-level generation and refinement of solution ideas, and 
(2)~an LLM \textbf{Instructor}, which translates these ideas into concrete code, drawing on domain-specific knowledge to address the complex needs of drug discovery tasks.


\paragraph{Problem Formulation.}
Following \citet{huang2024mlagentbench}, an ML programming task consists of three components: 
(1) a \textit{Task Description}, which specifies the objectives and constraints in natural language, 
(2) \textit{Starter Files}, which provide initial resources like datasets or code templates, and 
(3) \textit{Evaluator}, which is a performance metric function used to assess the output quality.



\paragraph{LLM Planner: Idea Space Management.}
Open-ended ML tasks in drug discovery can be approached by multiple strategies with no single deterministic solution, and single-agent systems risk missing promising alternatives~\cite{wang2024planning}. 
Additionally, LLMs sometimes make impractical suggestions if they lack specific domain expertise or rely on hallucinated information. 
To address these concerns, the Planner has two phases:
\begin{enumerate}[leftmargin=*]
    \item \emph{Idea Generation.} From the task description, the Planner derives \(K\) possible solution ideas.
    \item \emph{Exploration.} The Planner selects one idea and sends it to the Instructor for experimental evaluation. Based on success or failure reports, it revises the idea set, discarding those that underperform or are not feasible.
\end{enumerate}
The process repeats until a maximum iteration limit is reached, after which the highest-performing idea is submitted as the final solution.

\paragraph{LLM Instructor: Domain-specific Knowledge and Tool Preparation.}
\label{sub:instructor}
Drug discovery depends on specialized workflows, e.g., the correct handling of SMILES strings and tailored data preprocessing. 
When standard code-generation approaches ignore this domain requirements~\cite {huang2023hallucination,huang2024position}, the resulting errors are hard to debug.

Within \ours, the Instructor incorporates domain knowledge at every step of the coding process. It can execute standard ML actions (e.g., reading or editing scripts, running code; see Appendix~\ref{appendix:action}) and references a set of targeted documents to build or refine specialized tools. 
The Instructor then generates a performance report—if critical functionalities are absent, it returns a failure report instead.
Specifically, the Instructor relies on three curated types of documentation:
\begin{itemize}[leftmargin=*]
    \item \textbf{Raw Data Acquisition:} Methods for retrieving and preprocessing biological data.
    \item \textbf{Featurizing Biological Data:} Techniques for encoding molecules and proteins (e.g., fingerprints, graph-based representations).
    \item \textbf{Domain-Specific Models:} Pretrained foundation models such as ChemBERTa~\cite{chemberta} (small molecules) and ESM (Evolutionary Scale Modeling for protein sequence)~\cite{lin2022language}. 
\end{itemize}
Further details about these resources appear in Appendix~\ref{appendix:doc}. 
By explicitly integrating domain guidance into the coding workflow, \ours aims to reduce errors that arise from incomplete or incorrect handling of drug discovery subtleties.

\section{Experiment}

\subsection{Experimental Setup}

\paragraph{AI-solvable Drug Discovery Tasks.}
We propose three representative AI-solvable drug discovery tasks to validate the effectiveness of \ours. \textbf{ADMET} prediction forecasts pharmacokinetic properties (Absorption, Distribution, Metabolism, Excretion, and Toxicity) from a drug’s molecular structure, crucial for assessing a drug’s efficacy and safety~\cite{niu2024pharmabench,lu2022cot, lu2024uncertainty,chen2021data,chen2024uncertainty}. 
\textbf{High-throughput screening (HTS)} leverages ML models to predict assay outcomes based on molecular structure, improving the efficiency and reducing the cost of evaluating the biological activity of large chemical libraries~\cite{hts2021}. \textbf{Drug-target interaction (DTI)} prediction forecasts the binding affinity between drugs and proteins using compound structures and amino acid sequences, supporting virtual screening, drug repurposing, and side effect prediction~\cite{liu2024flexmol,zhang2021ddn2}.
All these problems are binary classification tasks. 

\vspace{-0.05in}

\paragraph{Dataset.}  
We select one dataset for each task: \textbf{PAMPA}~\cite{Siramshetty2021} for ADMET prediction, \textbf{DAVIS}~\cite{davis2011comprehensive} for DTI prediction, and \textbf{HIV}~\cite{wu2018moleculenet} for HTS. A detailed description of the datasets and the data splitting methods can be found in Appendix~\ref{appendix:dataset}.
\vspace{-0.05in}

\begin{table*}[t]
    \centering
    \resizebox{1\textwidth}{!}{
    \begin{tabular}{lcccccc}
        \toprule
        & \multicolumn{2}{c}{ADMET} & \multicolumn{2}{c}{HTS} & \multicolumn{2}{c}{DTI} \\
        \cmidrule(lr){2-3} \cmidrule(lr){4-5} \cmidrule(lr){6-7}
        Method & ROC-AUC ($\uparrow$) & Valid Rate ($\uparrow$) & ROC-AUC ($\uparrow$) & Valid Rate ($\uparrow$) & ROC-AUC ($\uparrow$) & Valid Rate ($\uparrow$) \\
        \midrule
        Human & 0.8173 & \textemdash{} & \textbf{0.8305} & \textemdash{} & 0.8940 & \textemdash{} \\
        CoT & 0.7599 & 62.5\% & 0.7524 & 50.0\% & N/A & 0.0\% \\
        React & 0.7385 & 87.5\% & 0.7653 & 75.0\% & 0.8530 & 50.0\% \\
        ResearchAgent & 0.7957 & 100.0\% & 0.7913 & 100.0\% & 0.8793 & 75.0\% \\
        DrugAgent@Top1 & 0.7667 & 100.0\% & 0.7919 & 100.0\% & 0.8950 & 87.5\% \\
        DrugAgent@Top3 & \textbf{0.8206} & 100.0\% & 0.8257 & 100.0\% & \textbf{0.8950} & 87.5\% \\
        \bottomrule
    \end{tabular}
    }
    \caption{ROC-AUC and Valid Rate for \textbf{PAMPA} (ADMET), \textbf{HIV} (HTS), and \textbf{DAVIS} (DTI) datasets. 
    }
    \label{tab:roc_auc_valid}
\end{table*}

\begin{table}[t]
    \centering
    \resizebox{\linewidth}{!}{%
    \begin{tabular}{lcc}
        \toprule
        Method & ROC-AUC ($\uparrow$) & Valid Rate ($\uparrow$) \\
        \midrule
        DrugAgent & 0.8950 & 87.5\% \\
        DrugAgent w/o Planner & 0.8845 & 87.5\% \\
        DrugAgent w/o Instructor & 0.8770 & 75.0\% \\
        \bottomrule
    \end{tabular}
    }
    \vspace{-0.7mm}
    \caption{%
    Ablation study on the DAVIS (DTI) task, demonstrating how removing the Planner or Instructor from \ours affects ROC-AUC and Valid Rate. 
    Results are averaged across runs.}
    \label{tab:ablation_dti}
\end{table}

\paragraph{Baselines.}
We use \texttt{GPT-4o-2024-08-06}~\cite{ChatGPT} to build all AI-based methods in our study. We compare \ours against three AI-based methods and one Human baseline. 
\textbf{CoT} (Chain of Thought) is a simple baseline where the agent generates a solution by breaking the problem into substeps~\cite{wei2022chain}. \textbf{ReAct} follows an interleaved reasoning and action approach, enabling interactive analysis and execution~\cite{yao2023react}. \textbf{ResearchAgent} is designed for ML tasks, maintaining a research plan and executing key actions such as file understanding, script editing, and task reflection~\cite{huang2024mlagentbench}. The \textbf{Human} baseline relies on model choices reported as effective in the literature and selected by experts, with details provided in Appendix~\ref{appendix:baseline}.

These baselines are compared with two variants of \ours: \textbf{DrugAgent@Top1}, where the agent selects the best solution based on validation results, and \textbf{DrugAgent@Top3}, where the agent submits the top three solutions based on validation results, and reports the best test set outcome. The detailed experimental settings for all agent frameworks are provided in Appendix~\ref{appendix:settings}.

\paragraph{Evaluation Metrics.}  
We conduct eight independent runs for each AI-based method. A submission is considered valid if (1) the generated code is free of bugs and, when executed, produces a submission file, (2) the submission file adheres to our format requirements, and (3) the performance does not fall more than 10\% below the human baseline. The average metric (ROC-AUC) across all valid submissions is reported. If all eight submissions are invalid, the results are marked as N/A.

\subsection{Quantitative Results}

Table \ref{tab:roc_auc_valid} reports the performance across all datasets. DrugAgent achieves the highest ROC-AUC and Valid Rate among all AI-based methods, performing comparably to baselines selected by human experts. Notably, it outperforms ReAct in the DTI task, achieving a relative improvement of 4.92\% in ROC-AUC. We also observe that DrugAgent@Top3 surpasses DrugAgent@Top1 in the ADMET and HTS tasks. This suggests that validation set performance does not always strongly correlate with test set performance, sometimes leading the agent to select a suboptimal idea for final submission. However, considering multiple submissions can help mitigate this problem.

Figure~\ref{tab:ablation_dti} highlights the importance of each agent in our framework, demonstrating that both the Planner and Instructor contribute significantly to overall performance. Additional qualitative analysis of their roles is provided in Appendix~\ref{appendix:ablation}.

\subsection{Case Study}
\paragraph{Comparing \ours with ReAct.}

We conduct a case study to compare our framework with ReAct (see Appendix~\ref{appendix:trace} for detailed traces and analysis). The results highlight our framework's effectiveness in diversifying ideas, accurately integrating domain knowledge, and learning from failures.

\begin{figure}[ht]
\centering
\includegraphics[width=\linewidth]{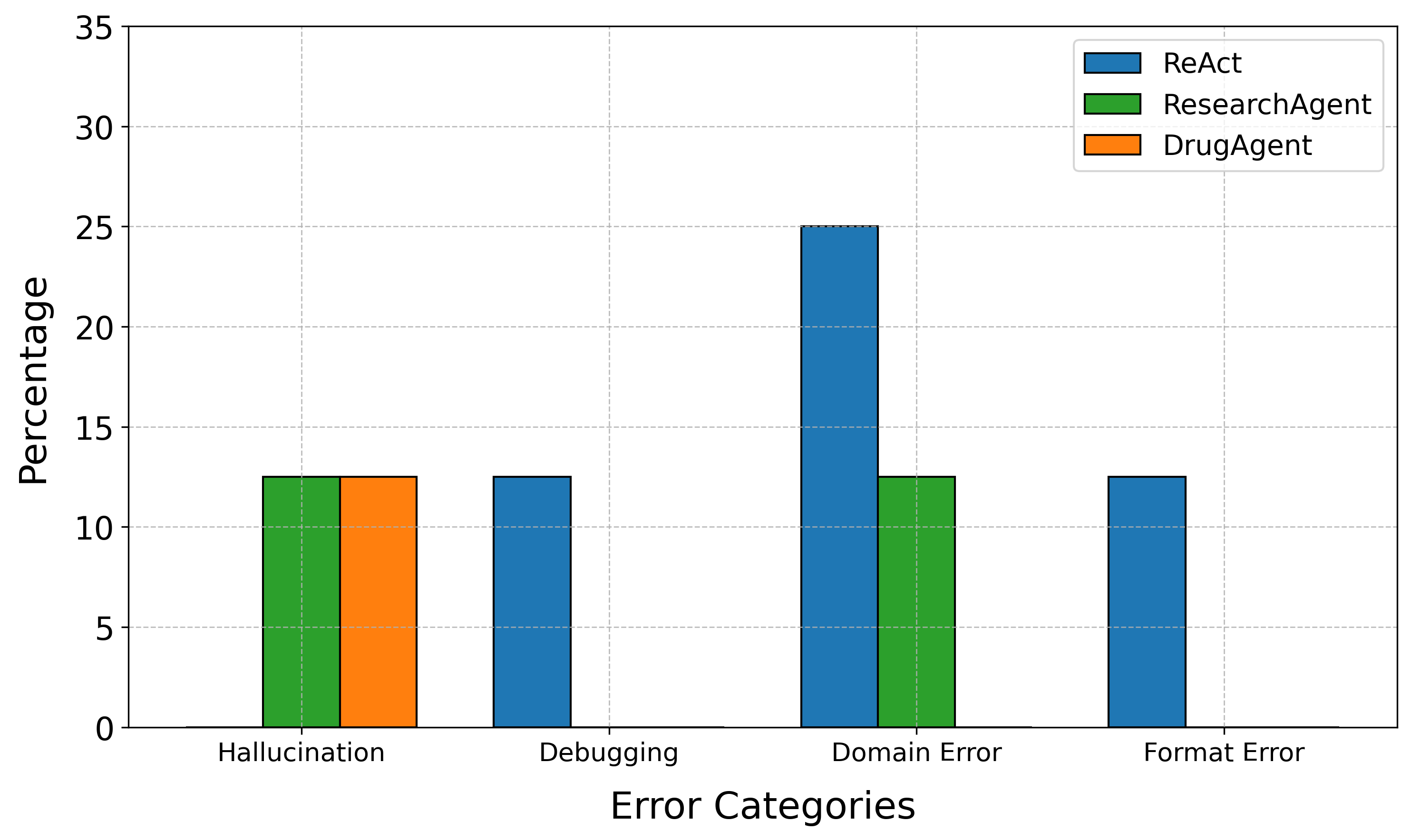}
\caption{Percentage of runs over \textbf{DAVIS} (DTI) dataset that falls into
different error modes.}
\vspace{-0.2in}
\label{fig:error_analysis}
\end{figure}

\paragraph{Trace Analysis.}
To further assess the agent’s reasoning and decision-making process, we analyze the traces of all runs for the DTI task and categorize the top four error types. A detailed description of each failure type is provided in Appendix~\ref{appendix:error}. Figure \ref{fig:error_analysis} illustrates that for general agent frameworks like ReAct and ResearchAgent, most errors occur due to poor performance caused by incorrect operations in steps requiring domain knowledge. In contrast, DrugAgent exhibits no errors in this category and achieves the lowest overall error rate, highlighting the effectiveness of our framework in utilizing domain knowledge.

\section{Conclusion}

In this paper, we have introduced \ours, a multi-agent framework that marks a significant advancement in leveraging large language models for automating critical aspects of drug discovery. Through case studies in three drug discovery tasks, \ours demonstrates remarkable improvements over general-purpose agent frameworks, such as ReAct and ResearchAgent. This can largely be attributed to the planner agent, which effectively generates and searches for ideas, and the instructor agent, which ensures reliable implementation by integrating a specialized toolset. Together, these agents enable \ours to bridge the gap between generalized AI capabilities and the nuanced demands of pharmaceutical research. We believe this work opens exciting new avenues for research and collaboration, pushing the boundaries of AI-driven drug discovery.


\section*{Limitations}
This study has several limitations. First, we evaluate the performance of DrugAgent on three case study tasks. However, these tasks are not sufficient for a comprehensive evaluation, and there is a need for more extensive benchmarks to assess machine learning programming tasks in drug discovery settings. Second, although DrugAgent can generate solutions comparable to human baselines, it is still limited to classic state-of-the-art baselines rather than the latest cutting-edge methods. Advancing agent capabilities in this domain will require significant research efforts. Third, the current documentation for DrugAgent is relatively basic and could be expanded in the future to cover additional aspects of the drug discovery process. Lastly, the agent framework has the potential to incorporate a 'human-in-the-loop' approach, which would enhance its usability for scientists working on real-world drug discovery tasks.

\section*{Ethics Statement}
We do not foresee any immediate ethical or societal concerns arising from our work. However, we acknowledge that, due to challenges like hallucination, the current version of DrugAgent is not yet ready for direct deployment in the drug discovery pipeline. For instance, errors such as fabricating results could lead to inaccurate predictions, which might waste resources in the wet lab verification process or misguide the drug discovery direction. As a result, further safety checks and human oversight are essential. 
Moreover, as AI agents advance, there is potential for them to replace human engineers in ML programming tasks within drug discovery. This highlights the need for human workers to learn how to effectively collaborate with the agent and understand its underlying implementation. By fostering this collaboration, AI can enhance and complement professional expertise rather than replace it.

\bibliography{acl}

\begin{thebibliography}{50}
\providecommand{\natexlab}[1]{#1}

\bibitem[{{CBDD Group}(2020)}]{pybiomed}
{CBDD Group}. 2020.
\newblock \href {https://codeload.github.com/gadsbyfly/PyBioMed/zip/refs/heads/master} {Pybiomed: A toolkit for calculating molecular descriptors and analyzing biological molecules}.
\newblock Accessed: February 13, 2025.

\bibitem[{Chen et~al.(2024{\natexlab{a}})Chen, Hu, Wang, Cao, Lin, Xu, Wu, Xiao, Sun et~al.}]{chen2024trialbench}
Jintai Chen, Yaojun Hu, Yue Wang, Xu~Cao, Miao Lin, Hongxia Xu, Jian Wu, Cao Xiao, Jimeng Sun, et~al. 2024{\natexlab{a}}.
\newblock Trialbench: Multi-modal artificial intelligence-ready clinical trial datasets.
\newblock \emph{arXiv:2407.00631}.

\bibitem[{Chen et~al.(2021)Chen, Lu, Wu, Clarke, Yu, Van~Eyk, Herrington, and Wang}]{chen2021data}
Lulu Chen, Yingzhou Lu, Chiung-Ting Wu, Robert Clarke, Guoqiang Yu, Jennifer~E Van~Eyk, David~M Herrington, and Yue Wang. 2021.
\newblock Data-driven detection of subtype-specific differentially expressed genes.
\newblock \emph{Scientific reports}, 11(1):332.

\bibitem[{Chen et~al.(2024{\natexlab{b}})Chen, Hao, and Van~Rechem}]{chen2024uncertainty}
Tianyi Chen, Nan Hao, and Capucine Van~Rechem. 2024{\natexlab{b}}.
\newblock \href {https://arxiv.org/abs/2401.03482} {Uncertainty quantification on clinical trial outcome prediction}.
\newblock \emph{Preprint}, arXiv:2401.03482.

\bibitem[{Davis et~al.(2011)Davis, Hunt, Herrgard, Ciceri, Wodicka, Pallares, Hocker, Treiber, and Zarrinkar}]{davis2011comprehensive}
Mindy~I Davis, Jeremy~P Hunt, Sanna Herrgard, Pietro Ciceri, Lisa~M Wodicka, Gabriel Pallares, Michael Hocker, Daniel~K Treiber, and Patrick~P Zarrinkar. 2011.
\newblock Comprehensive analysis of kinase inhibitor selectivity.
\newblock \emph{Nature biotechnology}, 29(11):1046--1051.

\bibitem[{Guo et~al.(2024)Guo, Chen, Wang, Chang, Pei, Chawla, Wiest, and Zhang}]{guo2024multiagent}
Taicheng Guo, Xiuying Chen, Yaqi Wang, Ruidi Chang, Shichao Pei, Nitesh~V. Chawla, Olaf Wiest, and Xiangliang Zhang. 2024.
\newblock \href {https://arxiv.org/abs/2402.01680} {Large language model based multi-agents: A survey of progress and challenges}.
\newblock \emph{Preprint}, arXiv:2402.01680.

\bibitem[{Huang et~al.(2021)Huang, Fu, Gao, Zhao, Roohani, Leskovec, Coley, Xiao, Sun, and Zitnik}]{Huang2021TherapeuticsDC}
Kexin Huang, Tianfan Fu, Wenhao Gao, Yue Zhao, Yusuf Roohani, Jure Leskovec, Connor~W. Coley, Cao Xiao, Jimeng Sun, and Marinka Zitnik. 2021.
\newblock Therapeutics data commons: Machine learning datasets and tasks for drug discovery and development.
\newblock \emph{Advances in Neural Information Processing Systems}.

\bibitem[{Huang et~al.(2022)Huang, Fu, Gao, Zhao, Roohani, Leskovec, Coley, Xiao, Sun, and Zitnik}]{huang2022artificial}
Kexin Huang, Tianfan Fu, Wenhao Gao, Yue Zhao, Yusuf Roohani, Jure Leskovec, Connor~W Coley, Cao Xiao, Jimeng Sun, and Marinka Zitnik. 2022.
\newblock Artificial intelligence foundation for therapeutic science.
\newblock \emph{Nature Chemical Biology}, 18:1033.

\bibitem[{Huang et~al.(2020{\natexlab{a}})Huang, Fu, Glass, Zitnik, Xiao, and Sun}]{huang2020deeppurpose}
Kexin Huang, Tianfan Fu, Lucas~M Glass, Marinka Zitnik, Cao Xiao, and Jimeng Sun. 2020{\natexlab{a}}.
\newblock {DeepPurpose}: a deep learning library for drug--target interaction prediction.
\newblock \emph{Bioinformatics}, 36(22-23):5545--5547.

\bibitem[{Huang et~al.(2020{\natexlab{b}})Huang, Xiao, Glass, and Sun}]{moltrans_2020}
Kexin Huang, Cao Xiao, Lucas~M Glass, and Jimeng Sun. 2020{\natexlab{b}}.
\newblock \href {https://doi.org/10.1093/bioinformatics/btaa880} {Moltrans: Molecular interaction transformer for drug–target interaction prediction}.
\newblock \emph{Bioinformatics}, 37(6):830–836.

\bibitem[{Huang et~al.(2023)Huang, Yu, Ma, Zhong, Feng, Wang, Chen, Peng, Feng, Qin, and Liu}]{huang2023hallucination}
Lei Huang, Weijiang Yu, Weitao Ma, Weihong Zhong, Zhangyin Feng, Haotian Wang, Qianglong Chen, Weihua Peng, Xiaocheng Feng, Bing Qin, and Ting Liu. 2023.
\newblock \href {https://doi.org/10.48550/arXiv.2311.05232} {A survey on hallucination in large language models: Principles, taxonomy, challenges, and open questions}.
\newblock \emph{arXiv preprint arXiv:2311.05232}.
\newblock Work in progress; 49 pages.

\bibitem[{Huang et~al.(2024{\natexlab{a}})Huang, Vora, Liang, and Leskovec}]{huang2024mlagentbench}
Qian Huang, Jian Vora, Percy Liang, and Jure Leskovec. 2024{\natexlab{a}}.
\newblock Mlagentbench: Evaluating language agents on machine learning experimentation.
\newblock In \emph{Thirty-eighth Conference on Neural Information Processing Systems}.

\bibitem[{Huang et~al.(2024{\natexlab{b}})Huang, Sun, Wang, Wu, Zhang, Li, Gao, Huang, Lyu, Zhang et~al.}]{huang2024position}
Yue Huang, Lichao Sun, Haoran Wang, Siyuan Wu, Qihui Zhang, Yuan Li, Chujie Gao, Yixin Huang, Wenhan Lyu, Yixuan Zhang, et~al. 2024{\natexlab{b}}.
\newblock Position: Trustllm: Trustworthiness in large language models.
\newblock In \emph{International Conference on Machine Learning}, pages 20166--20270. PMLR.

\bibitem[{Inaba et~al.(2023)Inaba, Kiyomaru, Cheng, and Kurohashi}]{inaba-etal-2023-multitool}
Tatsuro Inaba, Hirokazu Kiyomaru, Fei Cheng, and Sadao Kurohashi. 2023.
\newblock \href {https://aclanthology.org/2023.acl-short.130} {{M}ulti{T}ool-{C}o{T}: {GPT}-3 can use multiple external tools with chain of thought prompting}.
\newblock In \emph{Proceedings of the 61st Annual Meeting of the Association for Computational Linguistics (Volume 2: Short Papers)}, pages 1522--1532, Toronto, Canada. Association for Computational Linguistics.

\bibitem[{Jiang et~al.(2024)Jiang, Wang, Shen, Kim, and Kim}]{jiang2024survey}
Juyong Jiang, Fan Wang, Jiasi Shen, Sungju Kim, and Sunghun Kim. 2024.
\newblock \href {https://doi.org/10.48550/arXiv.2406.00515} {A survey on large language models for code generation}.
\newblock \emph{arXiv preprint arXiv:2406.00515}.

\bibitem[{Landrum(2023)}]{rdkit}
Greg Landrum. 2023.
\newblock \href {http://www.rdkit.org} {Rdkit: Open-source cheminformatics}.

\bibitem[{Li et~al.(2024{\natexlab{a}})Li, Yan, Pan, Luo, Ji, Ding, Xu, Liu, Dong, Lin, and Wang}]{li2024mmedagent}
Binxu Li, Tiankai Yan, Yuanting Pan, Jie Luo, Ruiyang Ji, Jiayuan Ding, Zhe Xu, Shilong Liu, Haoyu Dong, Zihao Lin, and Yixin Wang. 2024{\natexlab{a}}.
\newblock \href {https://doi.org/10.48550/arXiv.2407.02483} {Mmedagent: Learning to use medical tools with multi-modal agent}.
\newblock \emph{arXiv preprint arXiv:2407.02483}.
\newblock Accepted at EMNLP 2024.

\bibitem[{Li et~al.(2021)Li, Zhou, Hu, Fan, Zhang, Gu, and Karypis}]{dgllife}
Mufei Li, Jinjing Zhou, Jiajing Hu, Wenxuan Fan, Yangkang Zhang, Yaxin Gu, and George Karypis. 2021.
\newblock \href {https://doi.org/10.1021/acsomega.1c04017} {Dgl-lifesci: An open-source toolkit for deep learning on graphs in life science}.
\newblock \emph{ACS Omega}, 6(41):27233–27238.

\bibitem[{Li et~al.(2024{\natexlab{b}})Li, Zang, Ma, Guo, Zheng, Liu, Niu, Wang, Yang, Liu, Zhong, Zhou, Huang, and Zhang}]{li2024autokaggle}
Ziming Li, Qianbo Zang, David Ma, Jiawei Guo, Tuney Zheng, Minghao Liu, Xinyao Niu, Yue Wang, Jian Yang, Jiaheng Liu, Wanjun Zhong, Wangchunshu Zhou, Wenhao Huang, and Ge~Zhang. 2024{\natexlab{b}}.
\newblock \href {https://doi.org/10.48550/arXiv.2410.20424} {Autokaggle: A multi-agent framework for autonomous data science competitions}.
\newblock \emph{arXiv preprint arXiv:2410.20424}.

\bibitem[{Lin et~al.(2022)Lin, Akin, Rao, Hie, Zhu, Lu, dos Santos~Costa, Fazel-Zarandi, Sercu, Candido et~al.}]{lin2022language}
Zeming Lin, Halil Akin, Roshan Rao, Brian Hie, Zhongkai Zhu, Wenting Lu, Allan dos Santos~Costa, Maryam Fazel-Zarandi, Tom Sercu, Sal Candido, et~al. 2022.
\newblock Language models of protein sequences at the scale of evolution enable accurate structure prediction.
\newblock \emph{bioRxiv}.

\bibitem[{Liu et~al.(2025)Liu, Liu, Xu, Xia, and Li}]{liu2025sp_dti}
Sizhe Liu, Yuchen Liu, Haofeng Xu, Jun Xia, and Stan~Z. Li. 2025.
\newblock \href {https://doi.org/10.1093/bioinformatics/btaf011} {Sp-dti: Subpocket-informed transformer for drug-target interaction prediction}.
\newblock \emph{Bioinformatics}, btaf011.

\bibitem[{Liu et~al.(2024)Liu, Xia, Zhang, Liu, Liu, Du, Gao, Hu, Tan, Xiang, and Li}]{liu2024flexmol}
Sizhe Liu, Jun Xia, Lecheng Zhang, Yuchen Liu, Yue Liu, Wenjie Du, Zhangyang Gao, Bozhen Hu, Cheng Tan, Hongxin Xiang, and Stan~Z. Li. 2024.
\newblock Flexmol: A flexible toolkit for benchmarking molecular relational learning.
\newblock In \emph{Proceedings of the 38th Conference on Neural Information Processing Systems (NeurIPS)}.

\bibitem[{Lu et~al.(2024{\natexlab{a}})Lu, Lu, Lange, Foerster, Clune, and Ha}]{lu2024aiscientist}
Chris Lu, Cong Lu, Robert~Tjarko Lange, Jakob Foerster, Jeff Clune, and David Ha. 2024{\natexlab{a}}.
\newblock The {AI} {S}cientist: Towards fully automated open-ended scientific discovery.
\newblock \emph{arXiv preprint arXiv:2408.06292}.

\bibitem[{Lu et~al.(2024{\natexlab{b}})Lu, Chen, Hao, Van~Rechem, Chen, and Fu}]{lu2024uncertainty}
Yingzhou Lu, Tianyi Chen, Nan Hao, Capucine Van~Rechem, Jintai Chen, and Tianfan Fu. 2024{\natexlab{b}}.
\newblock Uncertainty quantification and interpretability for clinical trial approval prediction.
\newblock \emph{Health Data Science}, 4:0126.

\bibitem[{Lu et~al.(2022)Lu, Wu, Parker, Cheng, Saylor, Van~Eyk, Yu, Clarke, Herrington, and Wang}]{lu2022cot}
Yingzhou Lu, Chiung-Ting Wu, Sarah~J Parker, Zuolin Cheng, Georgia Saylor, Jennifer~E Van~Eyk, Guoqiang Yu, Robert Clarke, David~M Herrington, and Yue Wang. 2022.
\newblock {COT}: an efficient and accurate method for detecting marker genes among many subtypes.
\newblock \emph{Bioinformatics Advances}, 2(1):vbac037.

\bibitem[{Lála et~al.(2023)Lála, O'Donoghue, Shtedritski, Cox, Rodriques, and White}]{lala2023paperqa}
Jakub Lála, Odhran O'Donoghue, Aleksandar Shtedritski, Sam Cox, Samuel~G. Rodriques, and Andrew~D. White. 2023.
\newblock \href {https://doi.org/10.48550/arXiv.2312.07559} {Paperqa: Retrieval-augmented generative agent for scientific research}.
\newblock \emph{arXiv preprint arXiv:2312.07559}.

\bibitem[{M.~Bran et~al.(2024)M.~Bran, Cox, Schilter, Baldassari, White, and Schwaller}]{Bran}
Andres M.~Bran, Sam Cox, Oliver Schilter, Carlo Baldassari, Andrew~D. White, and Philippe Schwaller. 2024.
\newblock \href {https://doi.org/10.1038/s42256-024-00832-8} {Augmenting large language models with chemistry tools}.
\newblock \emph{Nature Machine Intelligence}, 6(5):525–535.

\bibitem[{Niu et~al.(2024)Niu, Xiao, Wu, Cai, Jiang, Jin, Wang, Yang, Kong, Jin, Yang, and Chen}]{niu2024pharmabench}
Zhangming Niu, Xianglu Xiao, Wenfan Wu, Qiwei Cai, Yinghui Jiang, Wangzhen Jin, Minhao Wang, Guojian Yang, Lingkang Kong, Xurui Jin, Guang Yang, and Hongming Chen. 2024.
\newblock \href {https://doi.org/10.1038/s41597-024-02072-y} {Pharmabench: Enhancing admet benchmarks with large language models}.
\newblock \emph{Scientific Data}, 11(985).

\bibitem[{Nowakowska(2023)}]{chemberta}
Sylwia Nowakowska. 2023.
\newblock \emph{Chemberta-2: Fine-tuning for molecule’s HIV replication inhibition prediction}.

\bibitem[{OpenAI(2024)}]{ChatGPT}
\href {https://doi.org/10.26434/chemrxiv-2023-b57vx} {OpenAI}. 2024.
\newblock \href {https://openai.com/research/gpt-4o} {Chatgpt: Gpt-4o-2024-08-06}.

\bibitem[{Pham et~al.(2021)Pham, Qiu, Zeng, Xie, and Zhang}]{hts2021}
Thai-Hoang Pham, Yue Qiu, Jucheng Zeng, Lei Xie, and Ping Zhang. 2021.
\newblock \href {https://doi.org/10.1038/s42256-020-00285-9} {A deep learning framework for high-throughput mechanism-driven phenotype compound screening and its application to covid-19 drug repurposing}.
\newblock \emph{Nature Machine Intelligence}, 3(3):247–257.

\bibitem[{Pushpakom et~al.(2019)Pushpakom, Iorio, Eyers, Escott, Hopper, Wells, Doig, Guilliams, Latimer, McNamee et~al.}]{Pushpakom2019}
Sudeep Pushpakom, Francesco Iorio, Patrick~A Eyers, K~Jane Escott, Shirley Hopper, Andrew Wells, Andrew Doig, Tim Guilliams, Joanna Latimer, Christine McNamee, et~al. 2019.
\newblock Drug repurposing: progress, challenges and recommendations.
\newblock \emph{Nature Reviews Drug Discovery}, 18(1):41--58.

\bibitem[{Qin et~al.(2023)Qin, Hu, Lin, Chen, Ding, Cui, Zeng, Huang, Xiao, Han, Fung, Su, Wang, Qian, Tian, Zhu, Liang, Shen, Xu, Zhang, Ye, Li, Tang, Yi, Zhu, Dai, Yan, Cong, Lu, Zhao, Huang, Yan, Han, Sun, Li, Phang, Yang, Wu, Ji, Liu, and Sun}]{qin2023tool}
Yujia Qin, Shengding Hu, Yankai Lin, Weize Chen, Ning Ding, Ganqu Cui, Zheni Zeng, Yufei Huang, Chaojun Xiao, Chi Han, Yi~Ren Fung, Yusheng Su, Huadong Wang, Cheng Qian, Runchu Tian, Kunlun Zhu, Shihao Liang, Xingyu Shen, Bokai Xu, Zhen Zhang, Yining Ye, Bowen Li, Ziwei Tang, Jing Yi, Yuzhang Zhu, Zhenning Dai, Lan Yan, Xin Cong, Yaxi Lu, Weilin Zhao, Yuxiang Huang, Junxi Yan, Xu~Han, Xian Sun, Dahai Li, Jason Phang, Cheng Yang, Tongshuang Wu, Heng Ji, Zhiyuan Liu, and Maosong Sun. 2023.
\newblock \href {https://arxiv.org/abs/2304.08354} {Tool learning with foundation models}.
\newblock \emph{Preprint}, arXiv:2304.08354.

\bibitem[{Ravuru et~al.(2024)Ravuru, Sakhinana, and Runkana}]{ravuru2024agentic}
Chidaksh Ravuru, Sagar~Srinivas Sakhinana, and Venkataramana Runkana. 2024.
\newblock \href {https://doi.org/10.48550/arXiv.2408.14484} {Agentic retrieval-augmented generation for time series analysis}.
\newblock In \emph{Proceedings of the Undergraduate Consortium at ACM SIGKDD Conference on Knowledge Discovery and Data Mining (KDD)}.

\bibitem[{Rives et~al.(2019)Rives, Meier, Sercu, Goyal, Lin, Liu, Guo, Ott, Zitnick, Ma, and Fergus}]{rives2019biological}
Alexander Rives, Joshua Meier, Tom Sercu, Siddharth Goyal, Zeming Lin, Jason Liu, Demi Guo, Myle Ott, C.~Lawrence Zitnick, Jerry Ma, and Rob Fergus. 2019.
\newblock \href {https://doi.org/10.1101/622803} {Biological structure and function emerge from scaling unsupervised learning to 250 million protein sequences}.
\newblock \emph{PNAS}.

\bibitem[{Schick et~al.(2023)Schick, Dwivedi-Yu, Dessì, Raileanu, Lomeli, Zettlemoyer, Cancedda, and Scialom}]{schick2023tool}
Timo Schick, Jane Dwivedi-Yu, Roberto Dessì, Roberta Raileanu, Maria Lomeli, Luke Zettlemoyer, Nicola Cancedda, and Thomas Scialom. 2023.
\newblock \href {https://arxiv.org/abs/2302.04761} {Toolformer: Language models can teach themselves to use tools}.
\newblock \emph{Preprint}, arXiv:2302.04761.

\bibitem[{Siramshetty et~al.(2021)Siramshetty, Shah et~al.}]{Siramshetty2021}
V.~B. Siramshetty, P.~Shah, et~al. 2021.
\newblock \href {https://doi.org/10.1177/24725552211017520} {Validating adme qsar models using marketed drugs}.
\newblock \emph{SLAS Discovery}, 26(10):1326--1336.

\bibitem[{Wang et~al.(2024{\natexlab{a}})Wang, Cassano, Wu, Bai, Song, Nath, Han, Hendryx, Yue, and Zhang}]{wang2024planning}
Evan Wang, Federico Cassano, Catherine Wu, Yunfeng Bai, Will Song, Vaskar Nath, Ziwen Han, Sean Hendryx, Summer Yue, and Hugh Zhang. 2024{\natexlab{a}}.
\newblock \href {https://doi.org/10.48550/arXiv.2409.03733} {Planning in natural language improves llm search for code generation}.
\newblock \emph{arXiv preprint arXiv:2409.03733}.

\bibitem[{Wang et~al.(2024{\natexlab{b}})Wang, Ma, Feng, Zhang, Yang, Zhang, Chen, Tang, Chen, Lin, and et~al.}]{Wang_agentsurvey}
Lei Wang, Chen Ma, Xueyang Feng, Zeyu Zhang, Hao Yang, Jingsen Zhang, Zhiyuan Chen, Jiakai Tang, Xu~Chen, Yankai Lin, and et~al. 2024{\natexlab{b}}.
\newblock \href {https://doi.org/10.1007/s11704-024-40231-1} {A survey on large language model based autonomous agents}.
\newblock \emph{Frontiers of Computer Science}, 18(6).

\bibitem[{Wang et~al.(2024{\natexlab{c}})Wang, Fu, Xu, Ma, Xu, Du, Gao, Wu, and Chen}]{wang2024twin}
Yue Wang, Tianfan Fu, Yinlong Xu, Zihan Ma, Hongxia Xu, Bang Du, Honghao Gao, Jian Wu, and Jintai Chen. 2024{\natexlab{c}}.
\newblock Twin-gpt: Digital twins for clinical trials via large language model.
\newblock \emph{ACM Transactions on Multimedia Computing, Communications and Applications}.

\bibitem[{WecoAI(2024)}]{aideml2024}
WecoAI. 2024.
\newblock \href {https://github.com/WecoAI/aideml} {Aide: The machine learning engineer agent}.

\bibitem[{Wei et~al.(2022)Wei, Wang, Schuurmans, Bosma, Ichter, Xia, Chi, Le, and Zhou}]{wei2022chain}
Jason Wei, Xuezhi Wang, Dale Schuurmans, Maarten Bosma, Brian Ichter, Fei Xia, Ed~Chi, Quoc Le, and Denny Zhou. 2022.
\newblock Chain-of-thought prompting elicits reasoning in large language models.
\newblock \emph{arXiv preprint arXiv:2201.11903}.

\bibitem[{Wolf et~al.(2020)Wolf, Debut, Sanh, Chaumond, Delangue, Moi, Cistac, Rault, Louf, Funtowicz, Davison, Shleifer, von Platen, Ma, Jernite, Plu, Xu, Scao, Gugger, Drame, Lhoest, and Rush}]{wolf-etal-2020-transformers}
Thomas Wolf, Lysandre Debut, Victor Sanh, Julien Chaumond, Clement Delangue, Anthony Moi, Pierric Cistac, Tim Rault, Rémi Louf, Morgan Funtowicz, Joe Davison, Sam Shleifer, Patrick von Platen, Clara Ma, Yacine Jernite, Julien Plu, Canwen Xu, Teven~Le Scao, Sylvain Gugger, Mariama Drame, Quentin Lhoest, and Alexander~M. Rush. 2020.
\newblock \href {https://www.aclweb.org/anthology/2020.emnlp-demos.6} {Transformers: State-of-the-art natural language processing}.
\newblock In \emph{Proceedings of the 2020 Conference on Empirical Methods in Natural Language Processing: System Demonstrations}, pages 38--45, Online. Association for Computational Linguistics.

\bibitem[{Wu et~al.(2018)Wu, Ramsundar, Feinberg, Gomes, Geniesse, Pappu, Leswing, and Pande}]{wu2018moleculenet}
Zhenqin Wu, Bharath Ramsundar, Evan~N Feinberg, Joseph Gomes, Caleb Geniesse, Ajay~S Pappu, Karl Leswing, and Vijay Pande. 2018.
\newblock Moleculenet: a benchmark for molecular machine learning.
\newblock \emph{Chemical science}, 9(2):513--530.

\bibitem[{Xia et~al.(2023)Xia, Zhang, Zhu, Liu, Gao, Hu, Tan, Zheng, Li, and Li}]{xia2023understanding}
Jun Xia, Lecheng Zhang, Xiao Zhu, Yue Liu, Zhangyang Gao, Bozhen Hu, Cheng Tan, Jiangbin Zheng, Siyuan Li, and Stan~Z. Li. 2023.
\newblock \href {https://arxiv.org/abs/2309.12440} {Understanding the limitations of deep models for molecular property prediction: Insights and solutions}.
\newblock In \emph{NeurIPS 2023}.
\newblock Last Modified: 03 Nov 2023.

\bibitem[{Yao et~al.(2023)Yao, Zhao, Yu, Du, Shafran, Narasimhan, and Cao}]{yao2023react}
Shunyu Yao, Jeffrey Zhao, Dian Yu, Nan Du, Izhak Shafran, Karthik Narasimhan, and Yuan Cao. 2023.
\newblock {ReAct}: Synergizing reasoning and acting in language models.
\newblock In \emph{International Conference on Learning Representations (ICLR)}.

\bibitem[{Yoon et~al.(2024)Yoon, Kim, and Oh}]{yoon2024designing}
Sion Yoon, Tae~Eun Kim, and Yoo~Jung Oh. 2024.
\newblock \href {https://doi.org/10.48550/arXiv.2406.19648} {Designing and evaluating multi-chatbot interface for human-ai communication: Preliminary findings from a persuasion task}.
\newblock \emph{arXiv preprint arXiv:2406.19648}.

\bibitem[{Yue et~al.(2024)Yue, Xing, Chen, and Fu}]{yue2024ct}
Ling Yue, Sixue Xing, Jintai Chen, and Tianfan Fu. 2024.
\newblock Clinicalagent: Clinical trial multi-agent with large language model-based reasoning.
\newblock \emph{arXiv preprint arXiv:2404.14777}.

\bibitem[{Zhang et~al.(2021)Zhang, Fu, Lu, Zhang, Clarke, Van~Eyk, Herrington, and Wang}]{zhang2021ddn2}
Bai Zhang, Yi~Fu, Yingzhou Lu, Zhen Zhang, Robert Clarke, Jennifer~E Van~Eyk, David~M Herrington, and Yue Wang. 2021.
\newblock {DDN}2.0: R and python packages for differential dependency network analysis of biological systems.
\newblock \emph{bioRxiv}, pages 2021--04.

\bibitem[{Öztürk et~al.(2018)Öztürk, Özgür, and Ozkirimli}]{deepdta}
Hakime Öztürk, Arzucan Özgür, and Elif Ozkirimli. 2018.
\newblock \href {https://doi.org/10.1093/bioinformatics/bty593} {Deepdta: Deep drug–target binding affinity prediction}.
\newblock \emph{Bioinformatics}, 34(17):i821–i829.

\end{thebibliography}

\appendix

\section{Related Work}
\label{appendix:related_work}


This section provides a more detailed overview of related work on LLM agents and their applications in ML programming and biomedical discovery.

\paragraph{LLM Agents}
An LLM agent is a system that uses large language models to interact with users or other systems, perform tasks, and make decisions autonomously. 
Empowered by LLMs, LLM agents have the capability to perform multi-step reasoning, planning, and action execution beyond static text generation~\cite{Wang_agentsurvey}. Previous works have equipped LLM agents with modules to dynamically interact with external tools, retrieve information, and adapt based on real-time feedback~\cite{schick2023tool, yoon2024designing, qin2023tool, ravuru2024agentic, lala2023paperqa}. This allows them to solve complex, evolving tasks such as code writing, long-term reasoning, and decision-making in various contexts~\cite{guo2024multiagent, jiang2024survey}. 
In this work, we tailor the LLM multi-agent framework to drug discovery tasks.

\paragraph{LLM for ML Programming}
Recent work has focused on accelerating traditionally manual research processes by automating ML programming.  AIDE acts as a data science agent, exploring a vast solution space and iteratively refining its approach to reach optimal solutions~\cite{aideml2024}. AutoKaggle introduces a specialized multi-agent framework for Kaggle data science competitions~\cite{li2024autokaggle}. AI-Scientist enables LLMs to conduct research autonomously, from idea generation to paper drafting, focusing on ML-related topics~\cite{lu2024aiscientist}. 
In parallel, benchmarks have been developed that provide a suite of 13 tasks to evaluate LLMs' capabilities in conducting ML programming~\cite{huang2024mlagentbench}.
However, existing works cannot handle domain-specific ML tasks requiring complex domain knowledge, e.g., AI-aided drug discovery. To address this, we design workflows to insert domain knowledge and call domain-specific tools automatically. 

\paragraph{LLM for Biomedical Discovery}
Many studies have highlighted the applications of LLMs in biomedical discovery, particularly when integrated with domain-specific tools. For instance, ChemCrow demonstrates the potential of LLM agents in organic synthesis, drug discovery, and material design~\cite{Bran}. Similarly, MMedAgent is a multimodal medical agent designed to handle complex language and multimodal tasks, demonstrating LLM versatility in medical applications~\cite{li2024mmedagent}. The multi-agent approach is exemplified by ClinicalAgent~\cite{yue2024ct}, which introduces a framework for clinical trial outcome prediction by decomposing it into subproblems, allowing individual agents to collaborate and generate a comprehensive outcome. Existing ML biomedical agents, however, generally lack the ML-specific expertise required to perform end-to-end programming.

\section{Action}
\label{appendix:action}
Below is a set of machine learning (ML)-related actions available to the instructor: List Files, Read File, Write File, Append File, Copy File, Inspect Script Lines, Undo Edit Script, Execute Script, Final Answer, Understand File, Edit Script, and Edit Script Segment. Since these actions are commonly used across general ML agents, we recommend referring to MLAgentBench~\cite{huang2024mlagentbench} for a detailed explanation of each action. 

\section{Documentation}
\label{appendix:doc}

\textbf{Raw Data Preprocessing}: We compiled documentation from the TDC library~\cite{Huang2021TherapeuticsDC}, which includes 66 AI/ML-ready datasets for drug discovery.

\textbf{Drug Preprocessing}: We documented seven molecular fingerprinting methods, two molecular graph construction methods, and one one-hot encoding method, using a combination of the TDC~\cite{Huang2021TherapeuticsDC}, DGL-LifeSci~\cite{dgllife}, and RDKit~\cite{rdkit} libraries.

\textbf{Protein Preprocessing}: We documented three protein fingerprinting methods and one one-hot encoding method, utilizing the PyBioMed~\cite{pybiomed} library.

\textbf{Domain-Specific Models}: We documented the ChemBERTa~\cite{chemberta} and ESM~\cite{rives2019biological} models, using the Transformers library~\cite{wolf-etal-2020-transformers}.

The complete documentation, along with the code for our framework, is available at \url{https://anonymous.4open.science/r/drugagent-5C42/}. It is important to note that this documentation can be easily extended based on specific needs and available resources.

\section{Dataset Description}
\label{appendix:dataset}

\begin{table*}[t]
  \centering
  \begin{tabularx}{\textwidth}{l|X|X|X}
    \toprule
     & \textbf{ADMET Prediction} & \textbf{HTS Prediction} & \textbf{DTI Prediction} \\
    \midrule
    \textbf{Type} & Single-instance prediction & Single-instance prediction & Multi-instance prediction \\
    \midrule
    \textbf{Input} & SMILES string & SMILES string & SMILES string and protein amino acid sequence \\
    \midrule
    \textbf{Impact} & Prevents clinical trial failures through early and accurate ADMET profiling & Reduces experimental screening costs by predicting assay outcomes & Reduces experimental screening needs by prioritizing drug candidates with high binding affinity \\
    \midrule
    \textbf{Dataset (Case Study)} & PAMPA~\cite{Siramshetty2021} & HIV~\cite{wu2018moleculenet} & DAVIS~\cite{davis2011comprehensive} \\
    \bottomrule
\end{tabularx}
\caption{Task overview: ADMET, HTS, and DTI. In this paper, we focus on small-molecule drugs, which constitute over 90\% of all approved drugs. Small molecules are represented as SMILES strings, a compact ASCII notation describing chemical structures.}
\label{tab:drug_discovery_overview}
\end{table*}

Table~\ref{tab:drug_discovery_overview} provides an overview of the selected drug discovery tasks and datasets used in our case study.  

\textbf{DAVIS}: This dataset contains 68 drugs and 379 proteins, with 2086, 3006, and 6011 samples allocated for training, validation, and testing, respectively. A detailed description of the dataset and preprocessing methods can be found in MolTrans~\cite{moltrans_2020}. The dataset is available at \url{https://github.com/kexinhuang12345/moltrans}.  

\textbf{PAMPA}: This dataset includes 1424 training samples, 203 validation samples, and 407 test samples. The data is split using the TDC random split strategy. More details can be found on the TDC website: \url{https://tdcommons.ai/single_pred_tasks/adme}.  

\textbf{HIV}: This dataset consists of 28,789 training samples, 4,113 validation samples, and 8,225 test samples. The split follows the TDC random split strategy. Further information is available on the TDC website: \url{https://tdcommons.ai/single_pred_tasks/hts}.

\section{Human Baseline}
\label{appendix:baseline}

Previous research~\cite{xia2023understanding} has shown that for ADMET and HTS tasks, tree-based models consistently outperform other approaches such as GCN, DNN, SVM, CNN, RNN, and MPNN. These models serve as a simple yet strong baseline that is difficult to beat. Therefore, we use a random forest model combined with Morgan fingerprinting as the human baseline for these two tasks.

For the DTI task, DeepDTA~\cite{deepdta}, which employs two CNN encoders for drug and protein representations, is a well-established deep learning baseline. It is widely adopted as a SOTA baseline in DTI studies~\cite{moltrans_2020, liu2024flexmol, liu2025sp_dti} and is considered the human baseline for this task.

\section{Settings}
\label{appendix:settings}
For all agent frameworks, we allow a maximum of 100 actions. For a detailed definition of an action, refer to Appendix~\ref{appendix:action} and the MLAgentBench~\cite{huang2024mlagentbench} paper. For the \textbf{ResearchAgent} baseline, we made the following adjustments to improve performance:
\begin{itemize}
    \item For the \texttt{understand\_file} action, we process only the first 3 blocks to save resources in case the file is too large (e.g., when understanding a CSV file).
    \item We also print error messages in the observation to assist the agent with debugging.
\end{itemize}

\section{Ablation Study}
\label{appendix:ablation}

\subsection{Without Instructor}

We found that although exploring multiple ideas improves the overall performance compared to the original ReAct framework, the results are still not satisfactory. The primary reason is that the model sometimes encodes molecules in an ineffective manner. Below is an example of code generated by the ReAct Agent that naively encodes a protein, leading to poor results despite a promising idea.

\begin{lstlisting}[language=Python, breaklines=true, columns=flexible]
def protein_to_features(protein_sequence):
    # Convert amino acid sequence into a feature vector of fixed length 1024
    features = np.zeros(1024, dtype=int)  # fixed length vector
    for i, c in enumerate(protein_sequence[:1024]):
        features[i] = ord(c)
    return features
\end{lstlisting}

\subsection{Without Planner}
We found that even when prompted to iteratively experiment with different models, the agent fails to sufficiently diversify its approach, often focusing on variations of similar ideas. For example, it may compare logistic regression with logistic regression incorporating feature engineering, which limits its ability to explore more optimal approaches.

\section{Comparing \ours with ReAct}
\label{appendix:trace}

\begin{figure*}[t]
\centering
\includegraphics[width=0.85\linewidth]{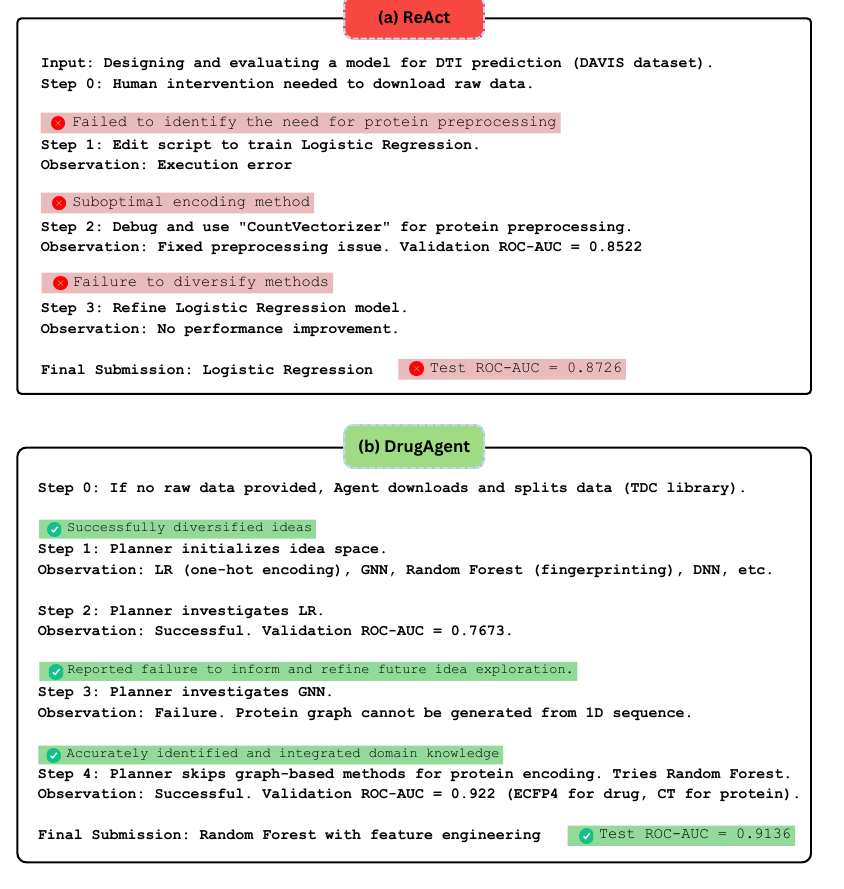}
  \caption{Comparison of ReAct and \ours on a DTI task. (a) ReAct, a general-purpose framework, delivers lower performance due to a lack of idea diversification and failure to recognize and incorporate domain knowledge. (b) \ours systematically explores a variety of approaches, successfully identifying optimal models and preprocessing methods to achieve strong performance.}
\label{fig:case_study}
\end{figure*}

To demonstrate the effectiveness of \ours, we conducted a case study on a DTI prediction task and compared its performance to ReAct, as illustrated in Fig.~\ref{fig:case_study}. This comparison underscores the challenges LLMs face in domain-specific tasks and highlights how \ours overcomes these limitations.

First, while ReAct~\cite{yao2023react} is prompted to iteratively select the best model, it lacks a high-level planning mechanism, instead focusing on implementing and refining a single approach. In contrast, \ours leverages a planner agent to diversify ideas and systematically identify the most effective approaches. 

Second, ReAct fails to recognize the need for preprocessing protein data early in the process, resulting in wasted time during debugging and the subsequent selection of suboptimal methods for molecular data encoding. By comparison, \ours correctly identifies substeps that require domain-specific tools and successfully integrates fingerprint-based encoding methods for biological data, delivering a bug-free model on the first attempt.

Third, ReAct is more prone to failure and struggles to recover from bad planning. \ours, on the other hand, learns from observations to guide future idea selection and avoids repeated failures. For instance, in the case study, \ours identifies the limitations of graph-based methods for protein encoding and avoids further exploration of those approaches. These findings highlight \ours's ability to automate domain-specific machine learning tasks while systematically selecting and refining the most effective approaches for the problem at hand.

\subsection{Sample Code by DrugAgent}

\begin{lstlisting}[language=Python, breaklines=true, columns=flexible]
import pandas as pd
import numpy as np
from sklearn.ensemble import RandomForestClassifier
from sklearn.metrics import roc_auc_score
from agent_tools import DrugFeaturizer
from agent_tools import ProteinFeaturizer

# Load train, validation, and test datasets
train_data = pd.read_csv("train.csv")
val_data = pd.read_csv("val.csv")
test_data = pd.read_csv("test.csv")

print(f"Loaded {len(train_data)} training samples, {len(val_data)} validation samples, and {len(test_data)} test samples.")

print(train_data.head())

X_train_drug = train_data['Drug']
X_train_protein = train_data['Protein']
y_train = train_data['Y']

X_val_drug = val_data['Drug']
X_val_protein = val_data['Protein']
y_val = val_data['Y']

# ************
# In this part of the code, write and train the model on the above dataset to perform the task.
# This part should populate the variable train_roc_auc and valid_roc_auc on the model selected
# ************

# Feature extraction
drug_featurizer = DrugFeaturizer(method='ECFP4')
protein_featurizer = ProteinFeaturizer(method='ct')

# Transformations
X_train_drug_features = drug_featurizer(X_train_drug.tolist())
X_train_protein_features = protein_featurizer(X_train_protein.tolist())

X_val_drug_features = drug_featurizer(X_val_drug.tolist())
X_val_protein_features = protein_featurizer(X_val_protein.tolist())

# Concatenate drug and protein features
X_train_features = np.concatenate([X_train_drug_features, X_train_protein_features], axis=1)
X_val_features = np.concatenate([X_val_drug_features, X_val_protein_features], axis=1)

# Train Random Forest model
rf_model = RandomForestClassifier(n_estimators=100, random_state=42)
rf_model.fit(X_train_features, y_train)

# Predict probabilities
train_preds = rf_model.predict_proba(X_train_features)[:, 1]
val_preds = rf_model.predict_proba(X_val_features)[:, 1]

# Compute ROC AUC scores
train_roc_auc = roc_auc_score(y_train, train_preds)
valid_roc_auc = roc_auc_score(y_val, val_preds)

# *********
# End of the main training module
# *********
print("Train ROC AUC Score: " + str(train_roc_auc))
print("Validation ROC AUC Score: " + str(valid_roc_auc))

X_test_drug = test_data['Drug']
X_test_protein = test_data['Protein']

# Transformations for test set
X_test_drug_features = drug_featurizer(X_test_drug.tolist())
X_test_protein_features = protein_featurizer(X_test_protein.tolist())
X_test_features = np.concatenate([X_test_drug_features, X_test_protein_features], axis=1)

# Replace with actual predictions
test_preds = rf_model.predict_proba(X_test_features)[:, 1]

test_data['Predicted'] = test_preds

output_file = "submission.csv" #do not change submission file name
test_data.to_csv(output_file, index=False)

print(f"Submission file saved to {output_file}.")
\end{lstlisting}

\section{Error Type}
\label{appendix:error}
\begin{enumerate}
    \item \textbf{Hallucination:} This occurs when the agent fabricates results or falsely claims progress, such as reporting a submission despite not making any edits to the training script.

    \item \textbf{Debugging:} The agent fails to resolve issues in its code modifications, such as mismatched tensor shapes.  
    
    \item \textbf{Domain Error:} Poor performance caused by incorrect operations in steps requiring domain knowledge (e.g., improper methods for fingerprinting drugs and proteins).

    \item \textbf{Format Error:} The agent altered the submission format, making it unrecognizable to the evaluator.
\end{enumerate}

\section{Code and Reproducibility}
The DrugAgent code is available at our anonymous repository: \url{https://anonymous.4open.science/r/drugagent-5C42/} and is under the MIT License.

\end{document}